\documentclass[11pt,a4paper]{article}
\usepackage[hyperref]{emnlp2018}
\usepackage{times}
\usepackage{latexsym}
\usepackage{graphicx}

\usepackage{url}

\aclfinalcopy 

\title{Learning to Generate Questions with Adaptive Copying Neural Networks}

\author{Xinyuan Lu \quad {and}\quad Yuhong Guo\\
School of Computer Science, Carleton University, Canada\\
  {\tt xinyuanlu@cmail.carleton.ca, yuhong.guo@carleton.ca} 
 } 

\date{}

\begin{document}
\maketitle

\begin{abstract}
Automatic question generation is 
an important problem in natural language processing. 
In this paper we propose 
a novel adaptive copying recurrent neural network model
to tackle the problem of 
question generation from sentences and paragraphs. 
The proposed model adds a copying mechanism component
onto a bidirectional LSTM architecture 
to generate more suitable questions adaptively from the input data. 
Our experimental results show the proposed model can outperform the state-of-the-art
question generation methods in terms of BLEU and ROUGE evaluation scores.  
  \end{abstract}
\section{Introduction}
Automatic question generation has recently 
received increasing attention in the natural language processing (NLP) research community.
The task is to generate proper questions from a given sentence or paragraph,
which has many applications in NLP, including generating questions 
for reading comprehension materials,
and developing dialog systems for building chat robots 
\cite{DBLP:journals/corr/MostafazadehMDZ16}. 
Moreover, as a reverse task of question answering, 
question generation can also be used to produce 
large scale question-answer pairs 
to assist question answering in NLP research.

Many previous works have used heuristic rule based methods to tackle
question generations \cite{Rus2010},
which have high requirements on the rule designs.  
The work in \cite{Heilman:2011:AFQ:2520603} proposed 
to compare generated candidate questions with a ranking algorithm and induce more suitable questions. 
The approach however depends on manually created features. 
A more recent work in \cite{DBLP:journals/corr/DuSC17} proposed a neural question generation model 
to automatically induce useful representations of input sentences or paragraphs and generate
suitable questions with LSTM networks. 
However given a limited size of annotated training data, sometimes this neural model could fail
to generate proper questions that are more suitable for the original inputs.

In this paper, we propose a new adaptive copying neural network (ACNN) 
model to tackle
the drawbacks of the previous works and generate proper questions.  
The proposed model exploits a bidirectional LSTM network with global attention mechanism 
to encode the sequential semantic information of the input sentence or paragraph.
When generating semantic questions with a LSTM decoder, 
it further incorporates a copying mechanism component 
to allow more suitable and natural words to be properly generated from
the source input sequence in a data adaptive manner. 
We conduct experiments on the most widely used Stanford Question Answering Dataset (SQuAD) \cite{DBLP:journals/corr/RajpurkarZLL16}. The empirical results show the proposed ACNN model can outperform 
the state-of-the-arts in terms of BLEU-$n$ and ROUGE-$L$ scores. 
%

%
\begin{figure*}[t]
\centering
\includegraphics[width=0.9\textwidth]{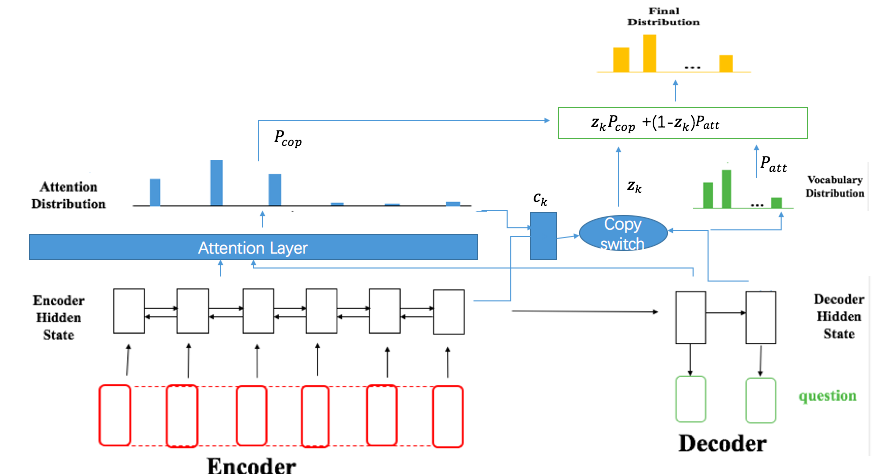}\\
\caption{The proposed Adaptive Copying Neural Network (ACNN). }
\label{fig1}
\end{figure*}
\section{Related Work}
{Question generation} (QG) 
has drawn a lot attention in the recent years. 
The previous work in \cite{article} applied minimal recursion semantics (MRS) to represent the meaning of sentences and then transfer MSR to questions. 
The work in \cite{Heilman:2011:AFQ:2520603} proposed an overgenerate-and-rank approach 
to generate and select high quality questions via ranking.
The authors of \cite{Chali:2015:TTG:2812180.2812181} focused on generating questions based on topics. 
The work in \cite{deep-questions-without-deep-understanding} used ontology crowd-sourcing to encode 
the original text in an ontology and align the question templates to select the most relevant ones.
The work in \cite{DBLP:journals/corr/SerbanGGACCB16} 
took the knowledge based information as input 
and generated questions based on it,
while  
\cite{DBLP:journals/corr/ZhouYWTBZ17} combined the answer position of the text. 
The recent work in \cite{DBLP:journals/corr/DuSC17} proposed a neural question generation model 
based on LSTMs which demonstrates good empirical results.

{Question answering} (QA) 
is a reverse task of QG. It shares similarities and sometimes mixed together with QG.
The works in \cite{DBLP:journals/corr/XiongZS16} and \cite{DBLP:journals/corr/ChenBM16a} 
used similar neural networks with attention mechanisms. 
The work of \cite{Song:2017:SAN:3018661.3018704} focused on retrieving non-factoid community questions as well as the lengths of the answers, while the work 
\cite{DBLP:journals/corr/YangHSC17} took reading comprehension as main tasks and question answering as auxiliary tasks. 
The authors of \cite{NIPS2016_6469} address QG and QA simultaneously to boost both of them.

\section{Model}
Given an input sentence or paragraph ${\bf x}$ which is a sequence of tokens 
$[x_1,\cdots, x_N]$, we aim to generate a natural question ${\bf y}=[y_1,\cdots,y_{|{\bf y}|}]$ from it.
Inspired by the work of
\cite{DBLP:journals/corr/DuSC17}, 
we use a bidirectional long short-term memory network (LSTM)
with global attention mechanism
to perform automatic question generation. 
We add a copying mechanism onto this neural model 
to incorporate
original input vocabulary information in the decoding phase to generate proper questions. 
The proposed ACNN model is demonstrated in Figure~\ref{fig1}.
This end-to-end learning model has two fundamental parts, attention based encoder and 
copying mechanism based decoder.

\subsection{Attention Based Encoder}

We use a bidirectional LSTM 
to encode the given sequence of tokens in the input sentence 
${\bf x}$. 
Let $\overrightarrow {{\bf h}_t}$ denote the hidden state at time step $t$ for the forward LSTM 
and  $\overleftarrow {{\bf h}_t}$ for the backward LSTM. 
The bidirectional LSTM produces the hidden states as follows:
\[
\overrightarrow {{\bf h}_t} = \overrightarrow{LSTM} (x_t, \overrightarrow{{\bf h}_{t-1}})
\]
\[
\overleftarrow {{\bf h}_t} = \overleftarrow{LSTM} (x_t, \overleftarrow{{\bf h}_{t+1}})
\]
By concatenating the hidden states
from both directions 
we have the following context dependent hidden representation at step $t$ 
${\bf h}_t=[\overrightarrow {{\bf h}_t};\overleftarrow {{\bf h}_t}]$. 

The attention based encoding of ${\bf x}$ at a decoding time step $k$ is then computed as
a weighted average of the representation vectors across ${\bf h}_t$,
\[
{\bf c}_k = \sum_{t=1}^N a_{k,t} {\bf h}_t.
\]
The attention weights $\{a_{k,t}\}$ are calculated using a softmax normalization 
\[
a_{k,t} = \frac{\exp(e_{k,t})}{\sum_{j=1}^N \exp(e_{k,j})},
\]
\[
e_{k,t} = \tanh({\bf d}_k^\top W_h {\bf h}_t) 
\]
where $W_h$ is the model parameter to be learned, and
${\bf d}_k$ is the hidden decoding state at time step $k$ 
which we will introduce below.

We also consider encoding
the truncated paragraph (with length $L$) that contains sentence ${\bf x}$ using the bidirectional 
encoding LSTM 
to replace the encoding of ${\bf x}$.

\subsection{Copying Mechanism Based Decoder}
The decoding process is to generate question ${\bf y}$ from the given sentence ${\bf x}$,
which is a probabilistic sequence prediction that can be factorized as:
\begin{eqnarray}
P({\bf y}|{\bf x}) = \prod_{k=1}^{|{\bf y}|}P(y_k | y<k, {\bf x})
\end{eqnarray}
This is also the empirical probability we need to maximize in the training process across 
all the annotated training instances.

We compute the local conditional word output probability 
$P(y_k | y<k, {\bf x})$ by integrating both a LSTM attention-based decoding component
and a copying mechanism component 
such that 
\begin{eqnarray}
&P(y_k | y<k, {\bf x}) 
\nonumber\\
= &z_k P_{cop}(y_k) + (1-z_t)P_{att}(y_k) 
\end{eqnarray}
The attention part
$P_{att}(y_k )$ generates words from
the common decoder vocabulary, 
and it is computed 
on the attention vector ${\bf c}_k$ and the hidden
state vector ${\bf d}_k$ from a decoding LSTM:
\[P_{att}(y_k)=softmax(W_y\tanh(W_k[{\bf d}_k;{\bf c}_k]))\] 
where $W_y$ and $W_k$ are model parameters.

The copying mechanism component $P_{cop}(y_k)$ 
generates (copies) words from the individual vocabulary of the source input sequence.
We compute it as
\begin{eqnarray}
P_{cop} (y_k) = softmax(\!V^\top\! (V [{\bf d}_k;\! {\bf c}_k]\!+\!b_1)\!+\!b_2)
\end{eqnarray}
when $y_k$ is from the unique word set of the source input sequence,
where $V, b_1$ and $b_2$ are model parameters.  
Such a copying mechanism can help 
incorporating words from the original data into the generated questions.

The combination weight $z_t$ is 
the switch for deciding generating the word from the vocabulary or 
copying it from the input sequence. 
We
calculate $z_t$ as follows:
\begin{eqnarray}
z_k = \sigma (W_d^\top {\bf d}_k + W_c^\top  {\bf c}_k + W_s^\top y_{k-1} + b)
\end{eqnarray}
where $W_d$, $W_c$, $W_s$ and $b$ are model parameters,
and $\sigma$ denotes a sigmoid function.
Hence $z_k$ functions as a selection gate that makes data adaptive selection
between the attention component and the copying component. 
\section{Experiments}

We conducted experiments on the widely used Stanford Question Answering Dataset (SQuAD) \cite{DBLP:journals/corr/RajpurkarZLL16}. 
We used the version released by \cite{DBLP:journals/corr/DuSC17}.
It was split into three parts -- training set, developing set and test set.
The training set contains 70,484 input-question pairs, 
the development set contains 10,570 input-question pairs, 
and the test set contains 11,877 input-question pairs. 

\begin{table*}[t]
\centering
\caption{The comparison results in terms of BLEU and ROUGE scores. 
The best scores in baselines and ACNN are highlighted using boldface.\\}
\label{tab1}
\begin{tabular}{l|cccc|c}
\hline
Model&BLEU-1&BLEU-2&BLEU-3&BLEU-4&ROUGE-$L$\\
\hline
Seq2Seq&31.34&13.79&7.36&4.26&29.75\\
Du-sent&\textbf{43.09}&\textbf{25.96}&\textbf{17.5}&\textbf{12.28}&\textbf{39.75}\\
Du-para&42.54&25.33&16.98&11.86&39.37\\
\hline
ACNN-sent &\textbf{44.78}&\textbf{26.83}&\textbf{18.72}&\textbf{13.97}&\textbf{41.08}\\
ACNN-para&44.37&26.15&18.02&13.49&40.57\\
\hline
\end{tabular}
\end{table*} 
\begin{table*}[t]
\centering
\caption{The results in terms of BLEU and ROUGE scores with different paragraph lengths. The best scores are highlighted using boldface.\\}
\label{tab2}
\begin{tabular}{l|cccc|c}
\hline
Model&BLEU-1&BLEU-2&BLEU-3&BLEU-4&ROUGE-$L$\\
\hline
ACNN-para-150&43.97&25.63&17.48&12.91&39.95\\
ACNN-para-120&44.22&25.94&17.80&13.26&40.33\\
ACNN-para-100&\textbf{44.37}&\textbf{26.15}&\textbf{18.02}&\textbf{13.49}&\textbf{40.57}\\
\hline
\end{tabular}
\end{table*}
%

\paragraph{Experimental setting}
The proposed model is built using Torch 7 on the OpenNMT system 
\cite{DBLP:journals/corr/KleinKDSR17}. 
We adopted the same setting as 
\cite{DBLP:journals/corr/DuSC17}.
We kept the most frequent 45K tokens as the encoder vocabulary and 
28K tokens as the decoder vocabulary. 
We used the word embeddings
released by \cite{pennington2014glove} as pre-training embeddings of the input words. 
%
We set the size of all LSTM hidden state vectors as 600 and the number of LSTM layers as 2. 
For the paragraph encoder, we set the length of paragraphs as 100. 
We use dropout technique in \cite{Srivastava:2014:DSW:2627435.2670313} with probability $p$ = 0.3. 
During testing, we set the beam search size to 3.
We used stochastic gradient descent(SGD) as optimization algorithm with initial learning rate $\alpha$ = 1.0 
and halve it when at epoch 8. To speed up training, we set mini-batch size to 64.

\paragraph{Comparison methods}
We compared the proposed ACNN model with the following methods on question generation.
(1) {\em Seq2Seq}: This model was proposed by \cite{NIPS2014_5346} 
which is a basic sequence to sequence transformation model. 
(2) {\em Du-sent}: This is the state-of-the art model developed in \cite{DBLP:journals/corr/DuSC17},
and we use it with the sentence level encoder and pre-trained word embeddings. 
(3) {\em Du-para}: This denotes the same model as above but with paragraph level encoder to be incorporated. 
For our proposed ACNN model, we also tested two versions, 
{\em ACNN-sent} with only sentence encoder and
{\em ACNN-para} with only paragraph encoder.

\paragraph{Evaluation metrics}
We used two types of commonly used evaluation metrics, BLEU-$n$ and ROUGE-$L$, to evaluate the testing results.
BLEU-$n$ (Bilingual Evaluation Understudy) \cite{Papineni:2002:BMA:1073083.1073135}
is a score that uses $n$ grams to calculate the correspondence 
between the machine generated output and the ground truth. 
ROUGE (Recall-Oriented Understudy for Gisting Evaluation) 
\cite{rouge-a-package-for-automatic-evaluation-of-summaries} 
measures the co-occurrences 
between the system-generated summary and the content in a human-generated summary. 
ROUGE-$L$ measures the co-occurrences of the longest common subsequence. 
The higher scores in these two metrics indicate better performance.

\subsection{Experimental Results}

The experimental results 
in terms of BLEU$1-4$ and ROUGE-$L$ scores 
for all the comparison methods
are reported in Table \ref{tab1}. 
The best results among the comparison baselines and the proposed ACNN variants 
are highlighted using boldface font separately.

We can see that among the comparison methods, 
\textit{Du-sent} produced the best performance 
in terms of all the evaluation metrics. 
\textit{Du-para}, though incorporated paragraph, produced slightly inferior performance.  
Both {\em Du-sent} and {\em Du-para} greatly outperform {\em Seq2Seq}.
Among the two variants of our proposed model, {\em ACNN-sent} slightly outperforms 
{\em ACNN-para} which uses paragraph as inputs.
This is consistent with the {\em Du-}methods and might be caused by the noise in the paragraphs.
Nevertheless, both variants of the proposed ACNN model consistently outperform the other comparison methods in terms of 
all the evaluation metrics. 
The comparison results between {\em ACNN-sent} and {\em Du-sent} validate the effectiveness of 
incorporating the copying mechanism into the bidirectional LSTM question generation model.

\subsection{Impact of Paragraph Length}
We also investigated the impact of the paragraph length on the performance 
of the proposed variant {\em ACNN-para}.  
We tested three different length values, 100 (the default value used above), 120 and 150. 
The comparison results are reported in 
Table \ref{tab2}.
We can see that when increasing the paragraph length, the test scores decrease.
This validates our analysis above on paragraph introducing noise:
Although longer paragraphs contain more contextual information, they include more irrelevant noisy information as well. 
\section{Conclusion and Future Work}
In this paper, we proposed an adaptive copying neural network (ACNN) model for question generation. 
We incorporated a copying mechanism component into a bidirectional LSTM model with global attention mechanism to improve its capacity on 
generating proper natural questions. 
We conducted experiments on the widely used 
{\em SQuAD} dataset and showed the proposed model outperforms the state-of-the-art method in the literature 
in terms of two types of evaluation metrics, 
BLEU-$n$ and ROUGE-$L$. 

In the future, we plan to extend the proposed model 
to address multi-task question generation on multi-documents with similar topics,
aiming to generate questions with similar copying mechanism that are consistent with human brain activities.  
The copying mechanism can also be expected to allow model adaptation across domains.

\bibliography{paperbib}
\bibliographystyle{acl_natbib_nourl}

\end{document}